\documentclass{article}

\usepackage{PRIMEarxiv}

\usepackage[utf8]{inputenc} 
\usepackage[T1]{fontenc}    
\usepackage{hyperref}       
\usepackage{url}            
\usepackage{booktabs}       
\usepackage{amsfonts}       
\usepackage{nicefrac}       
\usepackage{microtype}      
\usepackage{lipsum}
\usepackage{fancyhdr}       
\usepackage{graphicx}       
\graphicspath{{media/}}     
\usepackage{authblk}
\usepackage{graphicx,verbatim}
\usepackage{amsmath}   
\usepackage{amssymb}    
\usepackage{bm}         
\usepackage{mathtools}
\usepackage{booktabs}   
\usepackage{graphicx}
\usepackage{hyperref}
\usepackage{algorithm}
\usepackage{algorithmic}
\usepackage{multirow}
\usepackage{tabularx}
\usepackage{array}
\title{LEL: Lipschitz Continuity Constrained Ensemble Learning for Efficient EEG-Based Intra-subject Emotion Recognition}
\author[1,$\dagger$]{Shengyu Gong}
\author[1,$\dagger$]{Yueyang Li}
\author[1]{Zijian Kang}
\author[3]{Bo Chai}
\author[1,*]{Weiming Zeng}
\author[2]{Hongjie Yan}
\author[4]{Zhiguo Zhang}
\author[3]{Wai Ting Siok}
\author[3,*]{Nizhuan Wang}

\affil[1]{Lab of Digital Image and Intelligent Computation, Shanghai Maritime University, Shanghai 201306, China}
\affil[2]{Department of Neurology, Affiliated Lianyungang Hospital of Xuzhou Medical University, Lianyungang 222002, China}
\affil[3]{Department of Chinese and Bilingual Studies, The Hong Kong Polytechnic University, Hung Hom, Kowloon, Hong Kong Special Administrative Region, China}
\affil[4]{
    The Institute of Computing and Intelligence, Harbin Institute of Technology Shenzhen, Shenzhen 518000, China
}
\affil[$\dagger$]{Co-first authors}
\affil[*]{Correspondence: wangnizhuan1120@gmail.com; zengwm86@163.com}

\begin{document}
\maketitle
\begin{abstract}
Accurate and efficient recognition of emotional states is critical for human social functioning,  and impairments in this ability are associated with significant psychosocial difficulties. While electroencephalography (EEG) offers a powerful tool for objective emotion detection, existing EEG-based Emotion Recognition (EER) methods suffer from three key limitations: (1) insufficient model stability, (2) limited accuracy in processing high-dimensional nonlinear EEG signals, and (3) poor robustness against intra-subject variability and signal noise. To address these challenges, we introduce Lipschitz continuity-constrained Ensemble Learning (LEL), a novel framework that enhances EEG-based emotion recognition by enforcing Lipschitz continuity constraints on Transformer-based attention mechanisms, spectral extraction, and normalization modules. These heterogeneous constraints bound the global Lipschitz constant via function composition, ensures model stability, reduces sensitivity to signal variability and noise, and improves generalization capability. Additionally, LEL employs a learnable ensemble fusion strategy that optimally combines decisions from multiple heterogeneous classifiers to mitigate single-model bias and variance. Extensive experiments on three public benchmark datasets (EAV, FACED, and SEED) demonstrate superior performance, achieving average recognition accuracies of $74.25\% \pm 2.3$, $81.19\% \pm 2.8$, and $86.79\% \pm 1.9$, respectively. The official implementation codes are available at \url{https://github.com/NZWANG/LEL}.
\end{abstract}

\noindent\textbf{Note:} This work has been accepted by IEEE Sensors Journal.

\section{Introduction}
\label{sec:introduction}
Emotion recognition constitutes a foundational component of social cognition that facilitates interpersonal relationship formation and adaptive behavioral responses. Although sentiment analysis provides an efficient avenue for psychological assessment \cite{houssien2022review, li2022eeg} and has demonstrated utility in clinical monitoring \cite{hervas2023autism}, this capacity remains significantly compromised in various clinical populations. Specifically, distinct impairments characterize neurodevelopmental conditions such as Autism Spectrum Disorder (ASD) and Attention-Deficit/Hyperactivity Disorder (ADHD) \cite{li2025information, li2025mhnet, feng2024neural, kimhy2012emotion}, where emerging biomarker-based deep learning approaches offer alternative diagnostic pathways \cite{dogra2025development}. Similar deficits in processing emotional information are evident in psychiatric disorders including depression and schizophrenia \cite{zhang2024stanet, hirsch2018emotional, conner2021emotion}, with recent advances integrating EEG-based emotion recognition into BCI systems for therapeutic intervention \cite{zhang2024application}. These widespread dysfunctions \cite{joormann2016examining, cai2025mm, dong2025starformer} impose substantial personal and socioeconomic burdens, necessitating the development of objective recognition tools for precise clinical diagnostics and advanced computational applications. Recent systematic reviews have highlighted the rapid progress in constructing closed-loop EEG-based affective BCI (aBCI) systems, which require both accurate emotion recognition and effective neural regulation capabilities \mbox{\cite{Chen_2025}}.

EEG-based Emotion Recognition (EER) addresses this clinical need by providing objective assessment of emotional dysregulation, particularly in cases where communication barriers exist \cite{pillalamarri2025review}. Direct measurement of neurophysiological activity via EEG offers three distinct advantages: 1) high temporal resolution for capturing rapid emotional dynamics essential for mental health research, 2) non-invasive continuous monitoring capability for longitudinal studies, and 3) practical deployment feasibility through single-channel configurations \cite{10947211, Researchontwo-dimensional}. These collective capabilities enable early intervention and facilitate precision treatment strategies \cite{houssien2022review}. However, EEG signals are often noisy and susceptible to artifacts, which can limit the accuracy and robustness of emotion recognition models. Moreover, the interpretation of EEG data requires specialized expertise, potentially restricting its widespread clinical application.

\begin{figure*}
    \centering
    \includegraphics[width=1.0\textwidth]{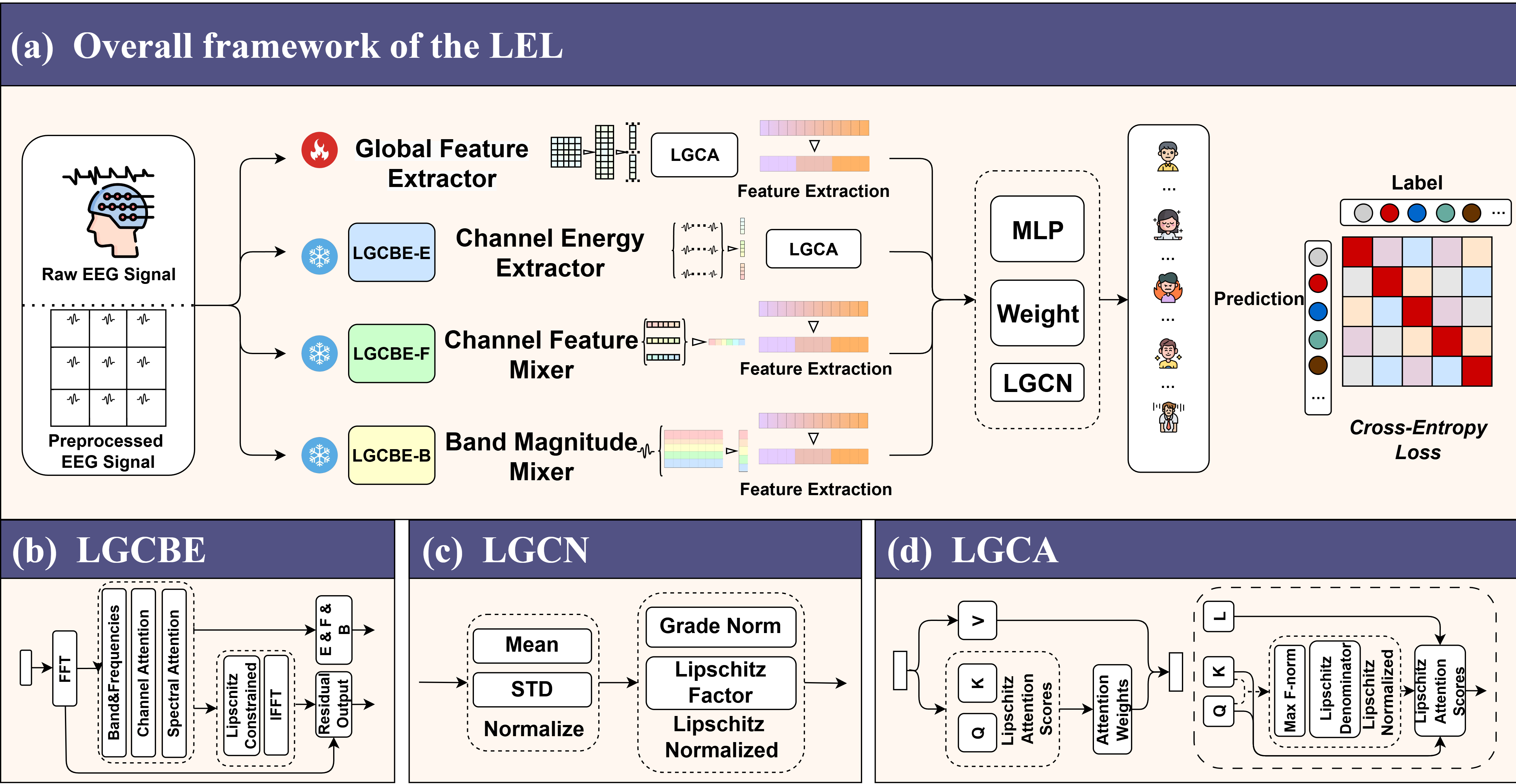}
    \caption{(a) LEL framework: Four lightweight branches fused for final emotion classification.
     (b) LGCBE: Lipschitz-constrained multi-band convolution and channel energy pooling.
     (c) LGCN: Lipschitz-constrained normalization.
     (d) LGCA: Lipschitz-constrained attention.}
    \label{fig:LEL}
\end{figure*}

The inherent complexity of emotion—shaped by neurobiological, interpersonal, cultural, and individual factors \cite{li2022eeg}—presents fundamental challenges. The absence of universal categorization frameworks and substantial expression heterogeneity in clinical populations \cite{hervas2023autism} necessitate precise quantification of internal states for diagnostics and personalized therapies \cite{li2024neural, ACase-basedChannel}. Recent EER advancements exemplify sophisticated solutions: FreqDGT \cite{li2025freqdgt} and the EEG emotion copilot \cite{chen2025eeg} leverage EEG's spatiotemporal dynamics to advance objective emotion modeling \cite{Spatial-temporalTransformers}. However, these advanced methods still face limitations in generalizability across different populations and contexts. The reliance on specific datasets and models may not capture the full spectrum of emotional experiences, and further research is needed to enhance the adaptability and scalability of these solutions.

Recent advances also explore graph-based approaches for modeling spatial dependencies in EEG signals. JAGP \mbox{\cite{9817639}} proposes a joint feature adaptation and graph adaptive label propagation framework for cross-subject emotion recognition, unifying domain-invariant feature learning with optimal graph learning. HDGNet \mbox{\cite{11262248}} introduces a hierarchical dynamic local-global graph representation learning method that captures both fine-grained local and coarse-grained global brain connectivity patterns. While these methods demonstrate the effectiveness of graph neural networks for EEG emotion recognition, they primarily focus on cross-subject or spatial modeling scenarios. In contrast, LEL addresses intra-subject emotion recognition with explicit stability guarantees through Lipschitz continuity constraints.

In this study, we propose the LEL (Lipschitz continuity-constrained Ensemble Learning) framework, whose core innovations are threefold:
\begin{enumerate}
\renewcommand{\labelenumi}{\arabic{enumi})}
	\item \textbf{Novel application of Lipschitz continuity constraints}: To solve model instability, limited generalization, and noise amplification, we systematically impose Lipschitz constraints on Transformer-based attention mechanisms, spectral extraction, and normalization modules (LGCBE, LGCA, LGCN) to enhance model stability and generalization when processing EEG signals.
	\item \textbf{Coupled constraint-ensemble architecture:} Unlike conventional EEG ensembles that post-hoc aggregate independently trained classifiers, LEL intrinsically couples learnable weights with Lipschitz constraints to optimize classifier fusion via $\mathrm{softmax}(w)$. The constraints ensure low-variance branch predictions, enabling the ensemble to safely exploit diverse temporal-spectral representations without amplifying individual branch instabilities, replacing traditional voting or uniform averaging methods.
	\item \textbf{Compositional robustness guarantees:} LEL achieves an accuracy of 74.25\% on EAV dataset, 81.19\% on FACED dataset, and 86.79\% on SEED dataset, demonstrating that modular Lipschitz bounds compose to ensure global stability in EEG-based emotion.
\end{enumerate}

The rest of this paper is organized as follows. Section \ref{sec:relatedwork} reviews recent progress and challenges in EER and Lipschitz continuity. Section \ref{sec:method} presents the LEL framework and its integration of Lipschitz constraints with ensemble learning. Section \ref{sec:experiments} evaluates the framework on multiple benchmark datasets. Section \mbox{\ref{sec: discussion}} discusses the experimental results and identifies directions for future work. Section \ref{sec: conclusion} concludes the paper's contribution.

\section{Related Work}
\label{sec:relatedwork}

\subsection{Recent Advances and Challenges in EER} 
Recent EER research focuses on improving accuracy \cite{geng2024deep, 10947211} and artifact robustness \cite{zhao2025adaptive}. Multimodal fusion integrates EEG with physiological signals like eye movement and heart rate for complementary information \cite{pillalamarri2025review, ZHANG2024121692}. Kang et al. employed hypergraph learning to jointly model EEG and peripheral signals, enhancing discriminability \cite{kang2025hypergraph}.
Another advance constructs brain network features to capture complex patterns and dynamic neural interactions \cite{ResearchProgressofEEG-Based}. 2D CNNs \cite{Researchontwo-dimensional}, spatial-temporal Transformers \cite{Spatial-temporalTransformers}, and case-based interpretable channel selection \cite{ACase-basedChannel} improve performance. Neural multimodal contrastive learning offers new perspectives for EEG analysis \cite{li2024neural}. A comprehensive systematic review by \mbox{\cite{KHARE2024102019}} identified persistent challenges including low SNR, artifact susceptibility, intra-individual variability, and cross-dataset generalization that limit current EER methods.
Yet four challenges remain. \textbf{(1) Low SNR}: Physiological and environmental noise obscure cortical signals, impeding feature extraction \cite{ResearchProgressofEEG-Based}. \textbf{(2) Artifact susceptibility}: Ocular, muscular, and motion artifacts introduce non-neural distortions \cite{imtiaz2025towards}. \textbf{(3) Intra-individual variability}: Cross-session neural activity fluctuations undermine consistency \cite{chaddad2023electroencephalography}. \textbf{(4) Cross-dataset generalization}: Models degrade dramatically when transferred across datasets or populations \cite{zhang2024mini, rashid2020current, ResearchonGCN-LSTM}.
Current mitigation relies on handcrafted denoising or dataset-specific tuning, limiting scalability. While graph-based methods such as JAGP \mbox{\cite{9817639}} and HDGNet \mbox{\cite{11262248}} have shown promise in modeling spatial dependencies for cross-subject scenarios, constraint-based ensemble frameworks are needed to automatically adapt to intra-individual variability while providing theoretical robustness guarantees for intra-subject recognition. Transformer-based approaches such as EmT have demonstrated promising results for cross-subject EEG emotion recognition by modeling temporal graph structures \mbox{\cite{10960695}}, these methods primarily address inter-subject generalization rather than intra-subject stability and efficiency, and lack explicit theoretical guarantees regarding robustness to signal noise and variability. 

\subsection{Lipschitz Continuity Constraints in Neural Networks} 

Lipschitz Continuity Constraints (LCCs) bound output change rates, providing formal stability guarantees. In machine learning, LCCs regularize models to mitigate noise amplification and enhance adversarial robustness \cite{Jia2023, jia2024aligning}. Existing research has demonstrated improved stability of Lipschitz-regularized graph neural networks under node perturbations \cite{Jia2023}, with subsequent work applying them to fairness-aware learning \cite{jia2024aligning}.

Integrating LCCs with attention mechanisms mitigates unbounded self-attention limitations, such as overconfidence and gradient instability \cite{Ye2023, Dasoulas2021}. Transformers have been regularized to suppress excessive attention on uncertain inputs \cite{Ye2023}, while Lipschitz normalization has been proposed for deep Graph Attention Networks to stabilize training \cite{Dasoulas2021}. Though primarily validated on vision and language tasks, this establishes a theoretical basis for stabilizing attention-based EEG models, which exhibit prevalent long-range dependencies and graph structures \cite{Spatial-temporalTransformers, ACase-basedChannel}.

Preliminary EEG studies have explored LCCs for controlling output variation across distributions \cite{fiorini2024eeg, zhong2020eeg}, but systematic integration with EER ensemble architectures remains unexplored. 

\section{Method}
\label{sec:method}
\subsection{Problem Definition \& Lipschitz Continuity}

Let $\mathbf{X} \in \mathbb{R}^{B \times C \times T}$ denote an EEG signal with $C$ channels and $T$ temporal samples. The emotion recognition task is formulated as estimating the posterior probability:
\begin{equation}
\hat{y} = \arg\max_{y \in \mathcal{Y}} P(Y = y \mid \mathbf{X}), \quad \mathcal{Y} = \{1, \dots, K\}.
\end{equation}

To enhance robustness against EEG artifacts and inter-subject variability, we enforce Lipschitz continuity on the learned feature mapping $f(\cdot)$ for each individual sample $\mathbf{x} \in \mathbb{R}^{B \times C \times T}$, requiring:
\begin{equation}
\|f(\mathbf{x}_1) - f(\mathbf{x}_2)\| \le L \|\mathbf{x}_1 - \mathbf{x}_2\|, \quad \forall \mathbf{x}_1, \mathbf{x}_2,
\end{equation}
We simulate Lipschitz boundaries by normalizing, clipping, and constraining both weights and attention weights. For the deep network $f = f_n \circ \dots \circ f_1$ comprising LEL's heterogeneous modules,the global Lipschitz constant satisfies $L_{\text{global}} \leq \prod_{i=1}^n L_i$ via the composition property,ensuring that local perturbation bounds at the module level deterministically constrain the final prediction's sensitivity. This constraint limits the rate of output change relative to input perturbations, effectively controlling gradient variations across the entire architecture, and mitigating the impact of noise and session drift on generalization stability. The proposed LEL showed in Fig.~\ref{fig:LEL} integrates LCCs into spectro-temporal extraction, attention, and normalization modules within a lightweight ensemble framework.

\subsection{Lipschitz Gradient-Constrained Band Extraction Component (LGCBE)}

The LGCBE component (Fig.~\ref{fig:LEL}(b)) performs frequency band decomposition and adaptive weighting under explicit Lipschitz constraints. It accepts an input tensor $\mathbf{X} \in \mathbb{R}^{B \times C \times T}$ and produces channel-wise energy $\mathbf{E}_b \in \mathbb{R}^{B \times C}$, channel features $\mathbf{F}_c \in \mathbb{R}^{B \times C \times F}$, and band features $\mathbf{F}_b \in \mathbb{R}^{B \times C \times F}$, where $B$ is batch size, $C$ is number of channels, $T$ is temporal length, $F$ is frequency dimension, and $|\mathcal{B}|=5$ denotes the number of frequency bands.

\paragraph{Frequency Band Decomposition} Let $F(\cdot)$ denote a frequency-band extraction operator that maps the full spectrum to band-specific subspaces. For each band $b \in \mathcal{B}$, $\mathbf{F}_b = F(\mathbf{F}_c; \text{low}_b, \text{high}_b) \in \mathbb{R}^{B \times C \times F_b}$, where $F_b = \text{high}_b - \text{low}_b$ is the band-specific frequency dimension.
 The input is transformed to the frequency domain via FFT, extracting energy from five predefined bands ($\delta,\theta,\alpha,\beta,\gamma$):
\begin{equation}
\mathbf{F}_c = \texttt{FFT}(\mathbf{X}, \text{dim}=-1) \in \mathbb{R}^{B \times C \times F_{\text{full}}}
\end{equation}
where $F_{\text{full}} = \lfloor T/2 \rfloor + 1$.

For each band $b \in \mathcal{B}$, the band-specific representation is:
\begin{equation}
\mathbf{F}_b = \mathbf{F}[\,:\,,\,:\,,\ell_b:u_b\,] \in \mathbb{R}^{B \times C \times F_b}, \quad \forall b \in \mathcal{B}
\end{equation}
\begin{equation}
\mathbf{E}_b = \sum_{k=1}^{F_b} |\mathbf{F}_b^{(k)}|^2 \in \mathbb{R}^{B \times C}, \quad \forall b \in \mathcal{B}
\end{equation}
where $[\ell_b:u_b]$ denotes slice operator along frequency axis, $\mathbf{F}_b^{(k)}$ is the $k$-th frequency slice of $\mathbf{F}_b$ and $\odot$ denotes element-wise multiplication.

\paragraph{Adaptive Weighting} Let $\mathbf{E}_b = [\mathbf{e}_{b,1}, \mathbf{e}_{b,2}, \dots, \mathbf{e}_{b,C}] \in \mathbb{R}^{B \times C}$ denotes the energy of channel $c$ in band $b$. Channel weights $\mathbf{W}_c$ and spectral weights $\mathbf{S}_s$ are computed via parallel attention branches:
\begin{equation}
\begin{split}
\mathbf{W}_c &= \sigma\!\left(\text{MLP}_c\!\left(\frac{1}{|\mathcal{B}|}\sum_{b\in\mathcal{B}}\mathbf{E}_b\right)\right) \in \mathbb{R}^{B \times C},\\
\mathbf{S}_s &= \text{Softmax}\!\left(\text{MLP}_s\!\left(\frac{1}{C}\sum_{c=1}^{C}\mathbf{E}_b\right)_{b\in\mathcal{B}}\right) \in \mathbb{R}^{B \times |\mathcal{B}|}
\end{split}
\end{equation}
where $\text{MLP}_c$ and $\text{MLP}_s$ share identical architecture, modeling channel and spectral importance, respectively.

\paragraph{Lipschitz Constraint} Lipschitz continuity is enforced through a unified Lipschitz constant $L_{\text{Lip}}$, which is provided as an external hyperparameter. This constant controls gradient scaling for both the spectral and channel branches, with spectral normalization applied to all linear layers:

\begin{equation}
\begin{split}
\tilde{\mathbf{S}}_s = \frac{L_{\text{Lip}} \mathbf{S}_s}{\|\mathbf{S}_s\|_2 + \epsilon}, \quad 
\tilde{\mathbf{W}}_c = \frac{L_{\text{Lip}} \mathbf{W}_c}{\|\mathbf{W}_c\|_2 + \epsilon}
\end{split}
\end{equation}
Here, $\tilde{\mathbf{S}}_s$ and $\tilde{\mathbf{W}}_c$ denote the Lipschitz-constrained updated values of the spectral and channel weights, respectively. $\epsilon$ is a small constant for numerical stability. The $\ell_2$-norm scaling achieves gradient control while constraining the output magnitudes within the specified Lipschitz bound.

\paragraph{Signal Reconstruction} The weighted frequency representation is transformed back to the time domain via inverse FFT. Prior to element-wise multiplication, $\tilde{\mathbf{W}}_c$ and $\tilde{\mathbf{S}}_s$ are broadcasted to match dimensions. The constrained weights $\tilde{\mathbf{S}}_s$ are applied per-band while $\tilde{\mathbf{W}}_c$ is shared across all bands:
\begin{equation}
\mathbf{X}_{\text{time}} = \texttt{IFFT}\bigl(\sum_{b \in \mathcal{B}} \mathbf{F}_b \odot \tilde{\mathbf{W}}_c \odot \tilde{\mathbf{S}}_s, \text{dim}=-1\bigr)
\end{equation}
The band features $\mathbf{F}_b \in \mathbb{R}^{B \times C \times F}$ are obtained by concatenating all band-specific representations $\{\mathbf{F}_b\}_{b \in \mathcal{B}}$ along the band dimension. The reconstructed signal is then normalized:
\begin{equation}
\begin{split}
\mathbf{X}_{\text{res}} &= \texttt{ResidualConnection}(\mathbf{X}, \mathbf{X}_{\text{time}}) \\
\mathbf{X}_{\text{out}} &= \texttt{LayerNorm}(\mathbf{X}_{\text{res}}) \in \mathbb{R}^{B \times C \times T}
\end{split}
\end{equation}

By bounding the spectral and channel weights via $L_{\text{Lip}}$, LGCBE ensures that frequency-domain perturbations propagate with limited amplification, contributing to the global stability bound through multiplicative composition with downstream LGCN and LGCA layers.

\subsection{Lipschitz Gradient-Constrained Normalization Component (LGCN)}

LGCN (Fig.~\ref{fig:LEL}(c)) extends LayerNorm by explicitly bounding the Lipschitz constant on the affine transform, which is achieved by a fixed Lipschitz bound $L_{\text{affine}}$, preventing gradient explosion and stabilizing training for shallow EEG feature layers. In detail, for input $\mathbf{Z} \in \mathbb{R}^{B \times D}$, the constrained normalization is implemented through fusion of standard normalization and spectrally-normalized affine as follow:
\begin{equation}
\begin{aligned}
\tilde{\mathbf{Z}} = \frac{\mathbf{Z} - \boldsymbol{\mu}}{\boldsymbol{\sigma} + \epsilon}, \quad
\mathbf{Z}_{\text{out}} = L_{\text{affine}} \cdot \frac{\gamma}{\|\gamma\|_2} \odot \tilde{\mathbf{Z}} + \beta
\end{aligned}
\end{equation}
where $\boldsymbol{\mu}, \boldsymbol{\sigma} \in \mathbb{R}^{D}$ are batch statistics, $\gamma, \beta \in \mathbb{R}^{D}$ are learnable parameters, and $\epsilon = 10^{-6}$.
This ensures the scaling satisfies $\|\mathbf{Z}_{\text{out}}(x_1) - \mathbf{Z}_{\text{out}}(x_2)\|_2 \le L_{\text{affine}} \|\tilde{\mathbf{Z}}(x_1) - \tilde{\mathbf{Z}}(x_2)\|_2$ for any $x_1, x_2$. This explicit bound acts as a stabilizing bottleneck that prevents gradient explosion and limits error accumulation when composed with LGCBE and LGCA modules, ensuring ensemble branches operate within bounded variance.

\subsection{Lipschitz Gradient-Constrained Attention Component (LGCA)}

Lipschitz Continuity enhances neural network robustness and generalization while stabilizing training~\cite{Li2024}. We design the LGCA module (Fig.~\ref{fig:LEL}(d)) to enhance model stability, robustness, and generalization ability by setting the Lipschitz constant to clip and constrain attention scores. Given input $\mathbf{X} \in \mathbb{R}^{B \times T \times D}$ ($B$: batch, $T$: seq length, $D$: embed dim), computation proceeds as:

\paragraph{Projection and Constrained Attention}
For each head $h \in \{1,\dots,H\}$, the input $\mathbf{X}$ is linearly projected into query, key, and value matrices ($\mathbf{Q}_h, \mathbf{K}_h, \mathbf{V}_h$) using learned weights $\mathbf{W}^{Q}_h, \mathbf{W}^{K}_h, \mathbf{W}^{V}_h \in \mathbb{R}^{D \times d_h}$ (where $d_h = D/H$) and corresponding biases. To enforce $L_{\text{att}}$-Lipschitz continuity, we clamp attention scores $\mathbf{S}_h$ to the range $[-c, c]$ before applying softmax, where $c = L_{\text{att}} \cdot \sqrt{d_h}$. This clamping operation bounds gradient magnitudes and produces the constrained attention weights $\tilde{\mathbf{A}}_h$. This truncate-and-normalize operation is defined as:
\begin{equation}
\begin{aligned}
\mathbf{S}_h &= \frac{\mathbf{Q}_h \mathbf{K}_h^\top}{\sqrt{d_h}} \in \mathbb{R}^{B \times T \times T}, \\
\tilde{\mathbf{A}}_h &= \operatorname{Softmax}\!\bigl(\text{clamp}(\mathbf{S}_h, -c, c)\bigr) \in \mathbb{R}^{B \times T \times T},
\end{aligned}
\end{equation}
where $L_{\text{att}}$ denotes the target Lipschitz constant for the attention module, and $\mathbf{K}_h^\top$ transposes the last two dimensions.

\paragraph{Output Projection}
Multi-head outputs are concatenated and projected:
\begin{equation}
\begin{aligned}
\mathbf{X}_{\text{att}} = \Bigl[\delta(\tilde{\mathbf{A}}_1)\mathbf{V}_1;\dots;\delta(\tilde{\mathbf{A}}_H)\mathbf{V}_H\Bigr]\mathbf{W}^{O} + \mathbf{b}^{O} \\
\in \mathbb{R}^{B \times T \times D},
\end{aligned}
\end{equation}
where $\mathbf{W}^{O} \in \mathbb{R}^{(H d_h) \times D}$, $\mathbf{b}^{O} \in \mathbb{R}^{D}$, and $\delta(\cdot)$ is dropout during training. This fuses multi-head outputs and restores dimension for residual connections. During training, $\mathbf{W}^{O}$ is spectrally normalized to $\|\mathbf{W}^{O}\|_2 \leq L_{\text{linear}}$ for complementary Lipschitz control, when composed with LGCN normalization,ensures the attention module's output sensitivity remains within the global Lipschitz product.

\subsection{Heterogeneous Branch Fusion Mechanism}

This mechanism adaptively fuses outputs from four heterogeneous branches via learnable weights $\boldsymbol{\alpha} \in \mathbb{R}^4$, where each branch $i$ produces a prediction vector $\hat{\mathbf{p}}_i \in \mathbb{R}^{B \times K}$ representing posterior probabilities for $K$ emotion classes. As shown in Fig.~\ref{fig:LEL}(a), each branch models temporal, spectral, spatial, and band-specific modalities respectively to maximize feature diversity. The final posterior is computed via Softmax weighting as follow:
\begin{equation}
\mathbf{w} = \text{Softmax}(\boldsymbol{\alpha}) \in \mathbb{R}^4, \quad
\hat{\mathbf{p}}_{\text{final}} = \sum_{i=1}^{4} w_i \hat{\mathbf{p}}_i
\end{equation}
where gradients flow only to $\boldsymbol{\alpha}$ to avoid overfitting.

\section{Experiments}
\label{sec:experiments}
\subsection{Datasets}
We evaluated LEL on three public EEG emotion recognition datasets with distinct features. 

\textbf{EAV Dataset}\cite{eav} includes synchronized EEG, audio, and video from 42 participants in natural conversations, with five emotion categories across 8,400 trials in active (speaking) and passive (listening) conditions (active trials used by default). 

\textbf{FACED Dataset}\cite{faced} comprises 32-channel EEG from 123 participants viewing 28 emotion-eliciting videos, covering nine categories with arousal/valence ratings as a large-scale benchmark. 

\textbf{SEED Dataset}\cite{seed} provides 62-channel EEG and eye-tracking from 15 participants watching film clips to elicit three affective states (positive, negative, neutral), with three sessions of 15 trials each. 

\subsection{Experimental Setup}

\subsubsection{Preprocessing \& Training \& Validation}

All EEG signals were preprocessed using the provided preprocessing scripts from the dataset or used in their readily preprocessed form as supplied by the dataset. To ensure statistical independence while mitigating temporal correlations, we employed an intra-subject protocol, randomly assigning each subject's independent trials to training (60\%), validation (10\%), and test (30\%) sets, with all segmented windows from any given trial retained within the same split. Given significant individual differences in neural signatures, electrode impedance, and head anatomy, we focused on personalized emotion recognition. For spectral feature extraction, we applied a consistent five-band decomposition strategy across all datasets: Delta (1--4 Hz), Theta (4--8 Hz), Alpha (8--13 Hz), Beta (13--30 Hz), and Gamma (30--50 Hz). Models were trained using the Adam optimizer (learning rate 3e-4, dropout rate=0.3), with hyperparameters tuned via grid search on pilot subjects. Performance was quantified using subject-averaged macro Accuracy (ACC) and F1-score (F1), following the same evaluation protocol as the baseline methods for fair comparison.

Preprocessing followed dataset-specific protocols with consistent physiological principles. For FACED, we used the official preprocessed version (bandpass 0.05–47 Hz, downsampled to 250 Hz, ICA-based ocular removal, common average re-referencing, and MAD-based bad electrode interpolation). For EAV, we used the provided preprocessed data (high-pass >0.5 Hz, 50 Hz notch filter, 20-second segmentation without ICA). For SEED, we followed the established pipeline (downsampled to 200 Hz, bandpass 0.3–50 Hz, manual EMG/EOG rejection, 1-second epochs). All datasets were spectrally decomposed into five bands (Delta: 1–4 Hz, Theta: 4–8 Hz, Alpha: 8–13 Hz, Beta: 13–30 Hz, Gamma: 30–50 Hz). Trial-wise partitioning ensured statistical independence and prevented data leakage.

\begin{table}[h]
\centering
\caption{Hyperparameter Configuration and Search Space}\label{tab:hyperparams}
\small
\begin{tabular}{lcc}
\hline
\textbf{Hyperparameter} & \textbf{Search Range} & \textbf{Selected Value} \\
\hline
Learning rate & \{1e-4, 3e-4, 1e-3\} & 3e-4 \\
Dropout rate & Fixed & 0.3 \\
Epoch & \{50, 100, 150\} & 100 \\
Batch Size & \{32, 64, 128\} & 128 \\
$L_s$ (Spectral) & \{0.5, 1.0, 1.5\} & 1.0 \\
$L_{att}$ (Attention) & \{0.5, 1.0, 1.5\} & 1.0 \\
$L_{affine}$ (Affine) & \{0.5, 1.0, 1.5\} & 1.0 \\
$L_{linear}$ (Linear) & \{0.5, 1.0, 1.5\} & 1.0 \\
\hline
\end{tabular}
\end{table}

\subsubsection{Validation Process}

We performed classification on multiple datasets (\ref{sec:classification}), analyze the effects of Lipschitz constraints on training (\ref{sec:lipanalysis}), validate ensemble superiority through ablation studies of component contributions (\ref{sec:ablation}), assess robustness to low-SNR signals in passively collected EEG from the EAV dataset (\ref{sec:passive}), evaluate real-time performance (\ref{sec:realtime}), and examine channel connectivity (\ref{sec:chord}).

\subsection{Classification Results on Diverse Datasets}
\label{sec:classification}

\begin{figure*}[h]
    \centering
    \includegraphics[width=0.9\linewidth]{bigtextconfusion.png}
    \caption{Confusion Matrices of LEL (Our Model) Across Three Datasets: EAV (a), FACED (b), and SEED (c).}
    \label{fig:Confusion Matrix}
\end{figure*}

We evaluated LEL on three emotion recognition datasets (EAV, FACED, SEED). For EAV, we compared with reported methods to follow the original protocol; for FACED and SEED, we benchmarked against other methods to test generalization across paradigms.

On EAV, LEL achieved 74.25\% accuracy and 73.94\% F1-score, outperforming AMERL's EEG benchmark (see Table \ref{tab:EAV result}, Fig. \ref{fig:Confusion Matrix}(a)). This showed EEG alone can surpass multimodal methods by modeling temporal dynamics and spatial topology. On FACED, LEL reached 81.19\% accuracy and 70.61\% F1-score, demonstrating strong handling of class imbalance (see Table \ref{tab:FACED result}, Fig.\ref{fig:Confusion Matrix}(b)), with the relatively lower F1 potentially attributable to the increased difficulty arising from the dataset's large number of classes and subjects. On SEED, LEL scored 86.79\% accuracy and 85.90\% F1-score, validating stability on long-duration signals (see Table \ref{tab:SEED result}, Fig.\ref{fig:Confusion Matrix}(c)). 

Consistent performance across multiple datasets proves that LEL captures core EEG emotion signals rather than overfitting, demonstrating robustness.

\begin{table}[H]
	\centering
	\caption{Performance Comparison of Different Methods on the EAV  Dataset}\label{tab:EAV result}
	\small
	\setlength{\tabcolsep}{3pt}
	\begin{tabularx}{\linewidth}{>{\centering\arraybackslash}p{1.0cm}>{\centering\arraybackslash}p{1.4cm}>{\centering\arraybackslash}p{2.0cm}>{\centering\arraybackslash}X>{\centering\arraybackslash}X>{\centering\arraybackslash}X}
		\hline
		\textbf{Dataset} & \textbf{Type} & \textbf{Method} & \textbf{ACC(\%)} & \textbf{F1(\%)} \\ \hline
		\multirow{6}{*}{EAV} & \multirow{3}{*}{Traditional} & SVM & 48.96 & 48.67 \\
            & & KNN & 50.87 & 47.68 \\
		& & Decision Tree & 45.25 & 45.28 \\ \cline{2-5}
		& \multirow{2}{*}{Others} & EAV-EEG~\cite{eav} & 59.51  & - \\
		& & AMERL~\cite{amerl} & 53.51 & - \\
		\cline{2-5}
		& \multirow{1}{*}{Ours} & \multirow{1}{*}{LEL} & \textbf{74.25 $\pm$ 2.3 ($p\approx0.003$)} & \textbf{73.94 $\pm$ 2.5 ($p\approx0.003$)} \\ \hline
	\end{tabularx}
\end{table}

\begin{table}[H]
	\centering
	\caption{Performance Comparison of Different Methods on the FACED Dataset}\label{tab:FACED result}
	\small
	\setlength{\tabcolsep}{3pt}
	\begin{tabularx}{\linewidth}{>{\centering\arraybackslash}p{1.0cm}>{\centering\arraybackslash}p{1.4cm}>{\centering\arraybackslash}p{2.0cm}>{\centering\arraybackslash}X>{\centering\arraybackslash}X>{\centering\arraybackslash}X}
		\hline
		\textbf{Dataset} & \textbf{Type} & \textbf{Method} & \textbf{ACC(\%)} & \textbf{F1(\%)} \\ \hline
		\multirow{10}{*}{FACED} & \multirow{3}{*}{Traditional} & SVM & 49.05 & 47.38 \\
            & & KNN & 45.44 & 44.63 \\
		& & Decision Tree & 41.46 & 41.71 \\ \cline{2-5}
		& \multirow{6}{*}{Others} & ACCNet~\cite{ACCNet} & 63.07  & 67.99 \\
		& & FC\_STGNN~\cite{fcstgnn} & 61.12 & 61.84 \\
            & & EEGNet~\cite{eegnet} & 61.55 & 60.61 \\
            & & LGGNet~\cite{lggnet} & 54.43 & 59.26 \\
            & & BF-GCN~\cite{bfgcn} & 55.23 & 63.67 \\
            & & FBCNet~\cite{fbcnet} & 63.14 & 62.93 \\
		\cline{2-5}
		& \multirow{1}{*}{Ours} & \multirow{1}{*}{LEL} & \textbf{81.19 $\pm$ 2.8 ($p<0.001$)} & \textbf{70.61 $\pm$ 2.4 ($p<0.001$)} \\ \hline
	\end{tabularx}
\end{table}

\begin{table}[H]
	\centering
	\caption{Performance Comparison of Different Methods on the SEED Dataset}\label{tab:SEED result}
	\small
	\setlength{\tabcolsep}{3pt}
	\begin{tabularx}{\linewidth}{>{\centering\arraybackslash}p{1.0cm}>{\centering\arraybackslash}p{1.4cm}>{\centering\arraybackslash}p{2.0cm}>{\centering\arraybackslash}X>{\centering\arraybackslash}X>{\centering\arraybackslash}X}
		\hline
		\textbf{Dataset} & \textbf{Type} & \textbf{Method} & \textbf{ACC(\%)} & \textbf{F1(\%)} \\ \hline
		\multirow{10}{*}{SEED} & \multirow{3}{*}{Traditional} & SVM & 66.84 & 66.73 \\
            & & KNN & 63.50 & 67.22 \\
		& & Decision Tree & 66.41 & 66.38 \\ \cline{2-5}
		& \multirow{6}{*}{Others} & ACCNet~\cite{ACCNet} & 80.50  &  79.43 \\
		& & FC\_STGNN~\cite{fcstgnn} & 69.50 & 68.22 \\
            & & EEGNet~\cite{eegnet} & 74.50 & 73.93 \\
            & & LGGNet~\cite{lggnet} & 72.24 & 71.96 \\
             & & BF-GCN~\cite{bfgcn} & 68.83 & 67.91 \\
             & & FBCNet~\cite{fbcnet} & 77.31 & 78.26 \\
		\cline{2-5}
		& \multirow{1}{*}{Ours} & \multirow{1}{*}{LEL} & \textbf{86.79 $\pm$ 1.9 ($p\approx0.002$)} & \textbf{85.90 $\pm$ 1.7 ($p\approx0.002$)} \\ \hline
	\end{tabularx}
\end{table}
  
\subsection{Effects of Lipschitz Continuity Constraints}
\label{sec:lipanalysis}
\subsubsection{Analysis of Training Process}

Lipschitz constraints embedded in LEL regularize optimization dynamics by bounding the sensitivity of the loss function to parameter perturbations, effectively preventing gradient explosion in deep networks processing high-dimensional EEG tensors. This boundedness significantly dampens oscillation amplitudes during training, whereas unconstrained models often exhibit erratic accuracy fluctuations and sudden loss spikes due to high-frequency artifacts such as muscle noise, eye blink potentials, or electrical interference. By ensuring that small input perturbations do not induce disproportionate gradient changes, the constraints maintain consistent optimization directions across epochs without aggressive learning rate scheduling or gradient clipping. This regularization is particularly critical for non-stationary, low-SNR EEG data, as it prevents over-reaction to transient noise bursts, reducing the risk of premature convergence to poor suboptima or overfitting to domain-specific artifacts rather than genuine neural patterns. Consequently, stabilized training dynamics enable reliable extraction of robust neurophysiological features, ensuring consistent generalization across diverse recording sessions and subject variations while avoiding the destabilizing volatility typical of unregularized architectures.

\subsubsection{Analysis with ROC Curve}

Under Lipschitz regularization, a model's decision boundary complexity and generalization are directly governed by Lipschitz constant, here is K. We designed a 3×3 parameter grid (epochs: 30/50/100; K=0.1/1.0/10.0) to investigate joint impacts on emotion classification (see Supplementary Material Fig.~S3). Results show K directly affects training: small epochs and K yield flat ROC curves, indicating limited learning capacity. As K increases appropriately, curves shift to the upper-left and steepen, accelerating complex feature learning and performance gains. However, excessive K causes fluctuations, signaling overfitting and poor generalization.

\subsubsection{Visualization Using t-SNE}

\begin{figure}[H]
    \centering
    \includegraphics[width=0.8\linewidth]{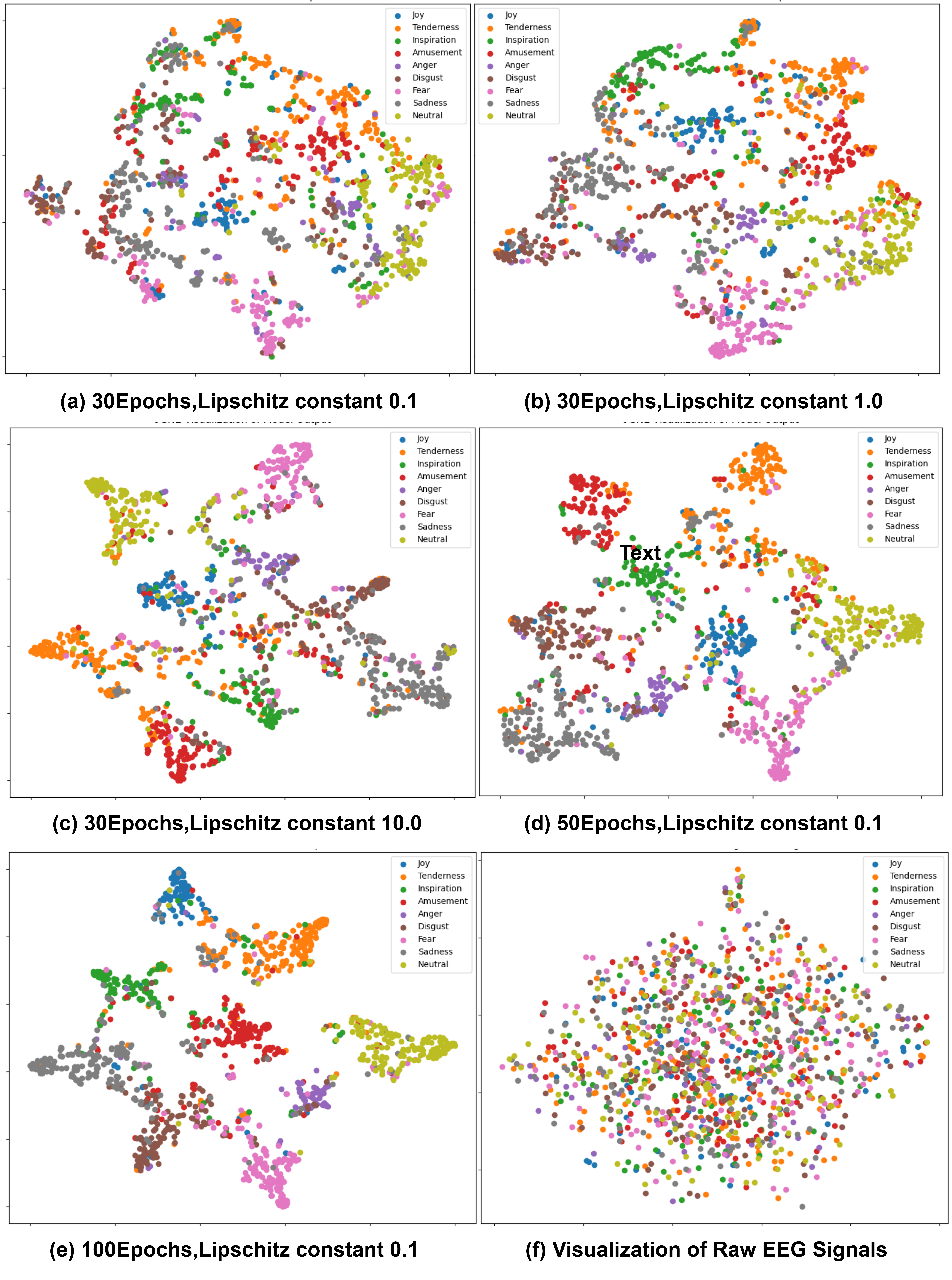}
    \caption{Six scatter plots comparing emotion classification across combinations of Lipschitz constants (0.1, 1.0, 10.0) and training epochs (30, 50, 100), with raw EEG in (f). Color-coded by nine emotion categories, showing parameter effects on feature space separation and clustering.}
    \label{fig:t-sne}
\end{figure}

We evaluate training epochs and Lipschitz constants on the nine-class FACED dataset using t-SNE visualizations (Fig.\ref{fig:t-sne}), testing epochs (30, 50, 100) and constants (0.1, 1.0, 10.0). Fig.\ref{fig:t-sne}(f) shows raw EEG signals with mixed categories, unclear boundaries, and no clusters. As training epochs increase (Fig.\ref{fig:t-sne}(d)), clustering improves progressively with clearer boundaries and more distinct clusters, approaching an ideal state at 100 epochs (Fig.\ref{fig:t-sne}(e)) with clear boundaries and tight, independent clusters. With sufficient training, smaller constants produce stable features and boundaries (Fig.\ref{fig:t-sne}(a)). As constants increase (Fig.\ref{fig:t-sne}(b) and Fig.\ref{fig:t-sne}(c)), clustering becomes tighter, accelerating detailed feature learning. Overall, smaller Lipschitz constants generate stable features with limited training, while moderate increases enable richer feature learning with sufficient training.

\subsection{Ablation Studies}
\label{sec:ablation}

Table~\ref{tab:Ablation Studies} shows ablation results across three datasets. Individual components exhibit dataset-dependent performance with significant variation, while the full model consistently outperforms all branches, achieving $74.25\%\pm2.3$ on EAV, $81.19\%\pm2.8$ on FACED, and $86.79\%\pm1.9$ on SEED. This performance gap demonstrates that Lipschitz constraints enable reliable ensemble diversity: by bounding each branch's variance ($L_{\mathrm{branch}}$), the constraints prevent dominant unstable experts from degrading fusion, allowing ensemble aggregation with learnable weights to dynamically balance contributions, mitigating overfitting and noise sensitivity. Thus, low inter-expert correlation validates that the framework exploits diverse, non-redundant features within bounded sensitivity for enhanced generalization.

\begin{table}[h]
    \centering
    \caption{Ablation study of different branches on three datasets.}
    \label{tab:Ablation Studies}
    \small
    \setlength{\tabcolsep}{3pt}
    \begin{tabularx}{\columnwidth}{
        >{\centering\arraybackslash}p{3.5cm}
        >{\centering\arraybackslash}p{1.0cm}
        >{\centering\arraybackslash}X
        >{\centering\arraybackslash}X}
        \hline
        \textbf{Model configuration} & \textbf{Dataset} & \textbf{ACC(\%)} & \textbf{F1(\%)} \\ \hline
        \multirow{3}{*}{Global Feature Extractor} 
        & EAV   & $69.56 \pm 2.8$    & $69.24 \pm 3.1$    \\
        & FACED & $61.95 \pm 3.2$    & $57.74 \pm 2.9$    \\
        & SEED  & $72.74 \pm 2.3$    & $72.68 \pm 2.4$     \\ \cline{2-4}
        \multirow{3}{*}{Channel Energy Extractor} 
        & EAV   & $68.23 \pm 2.5$    & $67.90 \pm 2.7$    \\
        & FACED & $71.67 \pm 3.5$    & $63.68 \pm 3.2$    \\
        & SEED  & $74.35 \pm 2.1$    & $74.43 \pm 2.5$    \\ \cline{2-4}
        \multirow{3}{*}{Channel Feature Mixer} 
        & EAV   & $61.01 \pm 3.2$    & $60.29 \pm 3.5$    \\
        & FACED & $76.20 \pm 3.9$    & $64.90 \pm 3.6$    \\
        & SEED  & $70.08 \pm 2.1$    & $70.19 \pm 2.2$  \\ \cline{2-4}
        \multirow{3}{*}{Band Magnitude Mixer} 
        & EAV   & $62.10 \pm 3.0$    & $61.72 \pm 3.3$    \\
        & FACED & $72.48 \pm 3.4$    & $59.08 \pm 3.3$    \\
        & SEED  & $71.94 \pm 2.2$    & $72.10 \pm 2.1$    \\ \cline{2-4}
        \multirow{3}{*}{Full model} 
        & EAV   & \textbf{74.25 $\pm$ 2.3} 
                & \textbf{73.94 $\pm$ 2.5} \\
        & FACED & \textbf{81.19 $\pm$ 2.8}                 
                & \textbf{70.61 $\pm$ 2.4} \\
        & SEED  & \textbf{86.79 $\pm$ 1.9}            
                & \textbf{85.90 $\pm$ 1.7} \\ \hline
    \end{tabularx}
\end{table}

\subsection{Validation on the EAV Passive Dataset}
\label{sec:passive}
The EAV dataset comprises active and passive signals. On active signals (recorded during standardized elicitation paradigms), LEL exhibits significant performance gains and robustness over baselines, maintaining stable discrimination despite physiological fluctuations. For more challenging passive signals (see Supplementary Material Fig.~S4) — characterized by high noise, low amplitudes, and weak label correlation — LEL maintains exceptional discriminative power, substantially outperforming other methods.
  
\subsection{Real-time Validation Test}
\label{sec:realtime}

\begin{figure}
    \centering
    \begin{minipage}[t]{0.49\textwidth}
        \centering
        \includegraphics[width=1.0\linewidth]{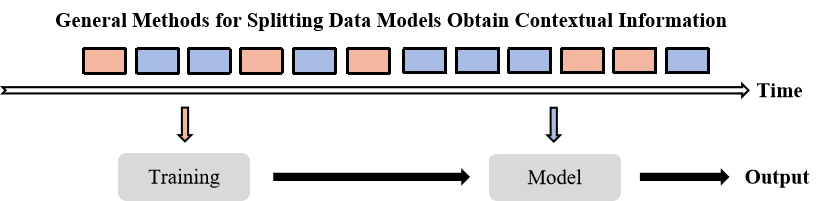}
        \caption{Partitioning with contextual information.}
        \label{fig:with contextual}
    \end{minipage}
    \hfill
    \begin{minipage}[t]{0.49\textwidth}
        \centering
        \includegraphics[width=1.0\linewidth]{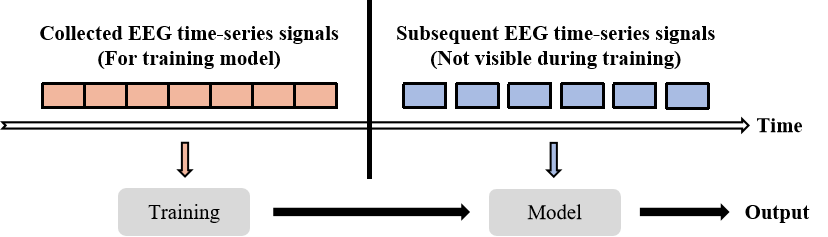}
        \caption{Partitioning without contextual information.}
        \label{fig:without contextual}
    \end{minipage}
\end{figure}

Real-time validation is more demanding than traditional methods. While the latter optimize accuracy using full context, real-time systems work only with current signals (Fig.~\ref{fig:with contextual} and~\ref{fig:without contextual}), requiring stronger generalization and robustness. As shown in Fig.~\ref{fig:realtime}), LEL's temporal modeling effectively extracts EEG features, suppresses noise, and maintains stable performance across complex settings, proving its practical reliability.

\begin{figure*}[h]
    \centering
    \includegraphics[width=0.9\linewidth]{realtime.png}
    \caption{Real-time validation test on EAV (a), FACED (b) and SEED (c)}
    \label{fig:realtime}
\end{figure*}

\subsection{Correlation Between EEG Channels and Emotions}
\label{sec:chord}

Chord diagrams (Fig.~S5) visualize inter-channel relationships for nine FACED emotions, with color intensity representing fused attention weights and signal correlations. These reveal emotion-specific signatures consistent with frontal asymmetry theory: positive emotions (Amusement, Joy) exhibit left-lateralized frontal connectivity (F3, Fp1) for approach motivation, whereas high-arousal negative emotions (Anger, Fear) display right-lateralized frontal coupling (F4, F8) reflecting defensive vigilance. Tenderness and Disgust show temporal lobe involvement (T3--T6) suggesting social-affective processing, while Sadness demonstrates symmetric frontal-parietal coupling indicative of rumination. These connectivity patterns function as neural fingerprints—where topology encodes emotion category and connection strength tracks arousal—providing interpretable biomarkers for affective computing.

\section{Discussion}
\label{sec: discussion}

Experimental results reveal limitations and directions for improvement of LEL. Ablation studies show significant dataset-dependent performance variations across branches; for instance, the Channel Feature Mixer achieves only 61.01\% accuracy on EAV but 76.20\% on FACED, suggesting the need for data-driven dynamic branch selection mechanisms rather than the fixed four-branch architecture. t-SNE visualizations indicate that smaller Lipschitz constants generate stable features yet require more training epochs for convergence, whereas excessively large constants lead to overfitting, necessitating careful tuning of constraint strength for specific datasets in practical deployment. Moreover, the currently designed framework only focuses on intra-subject emotion recognition and still faces the performance degradation in cross-subject scenarios due to EEG non-stationarity and individual differences in our cross-subject experiments setting. This implies that our future work could integrate domain adaptation techniques for cross-subject transfer, employ meta-learning strategies for rapid adaptation to new sessions with minimal labeled data, and extend Lipschitz normalization modules to incorporate session-adaptive bounds that dynamically adjust continuity constants based on real-time domain shift estimation, thereby extending LEL to more clinically practical applications.

\section{Conclusion}
\label{sec: conclusion}

We propose LEL, an EEG-based intra-subject emotion recognition framework that enforces Lipschitz constraints across frequency extraction, attention, and normalization modules to enhance training stability and noise robustness. By synergistically regularizing and fusing complementary representations (temporal, spectral, spatial, and band-specific) through learnable ensemble weights, LEL achieves average recognition accuracies of 74.25\%, 81.19\%, and 86.79\% on the EAV, FACED, and SEED datasets, respectively, maintaining robust performance even under low-SNR conditions. Ablation studies validate that Lipschitz constraints effectively control branch variance and improve ensemble fusion stability, while t-SNE and ROC analyses further confirm the regulatory effects of constraint strength on feature separation and model convergence. These results demonstrate that modular Lipschitz bounds achieve global stability guarantees through function composition.

\renewcommand{\thefigure}{S\arabic{figure}}
\setcounter{figure}{0}
\section{Supplementary Material}
\subsection{Supplementary Figure S1 and S2 for Analysis of Training Process}
\begin{figure}[H]
    \centering
    \includegraphics[width=0.6\textwidth]{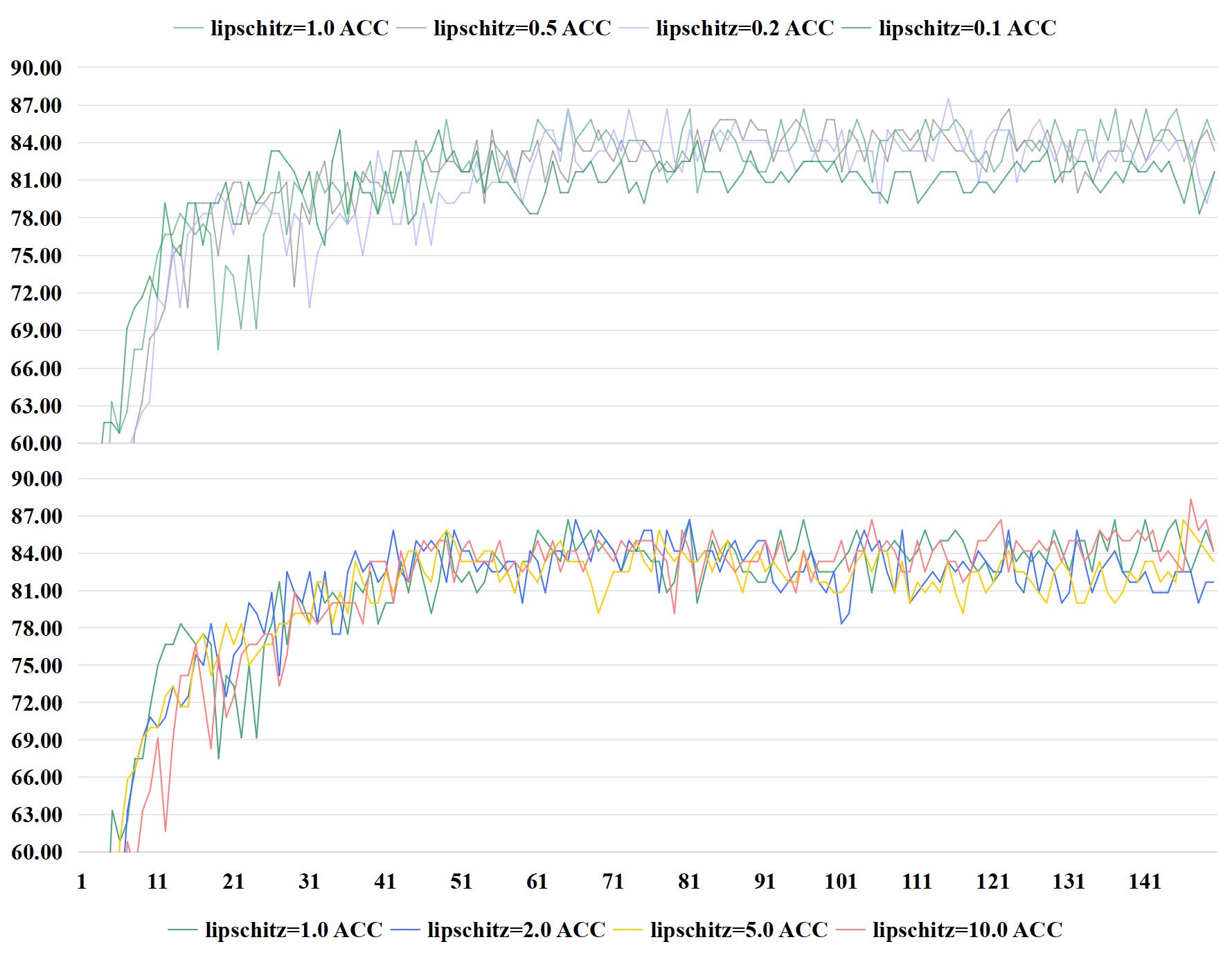}
    \caption{EAV Sub2 with different Lipschitz constant.}
    \label{fig:eav-lc1}
\end{figure}

\begin{figure}[H]
    \centering
    \includegraphics[width=0.6\textwidth]{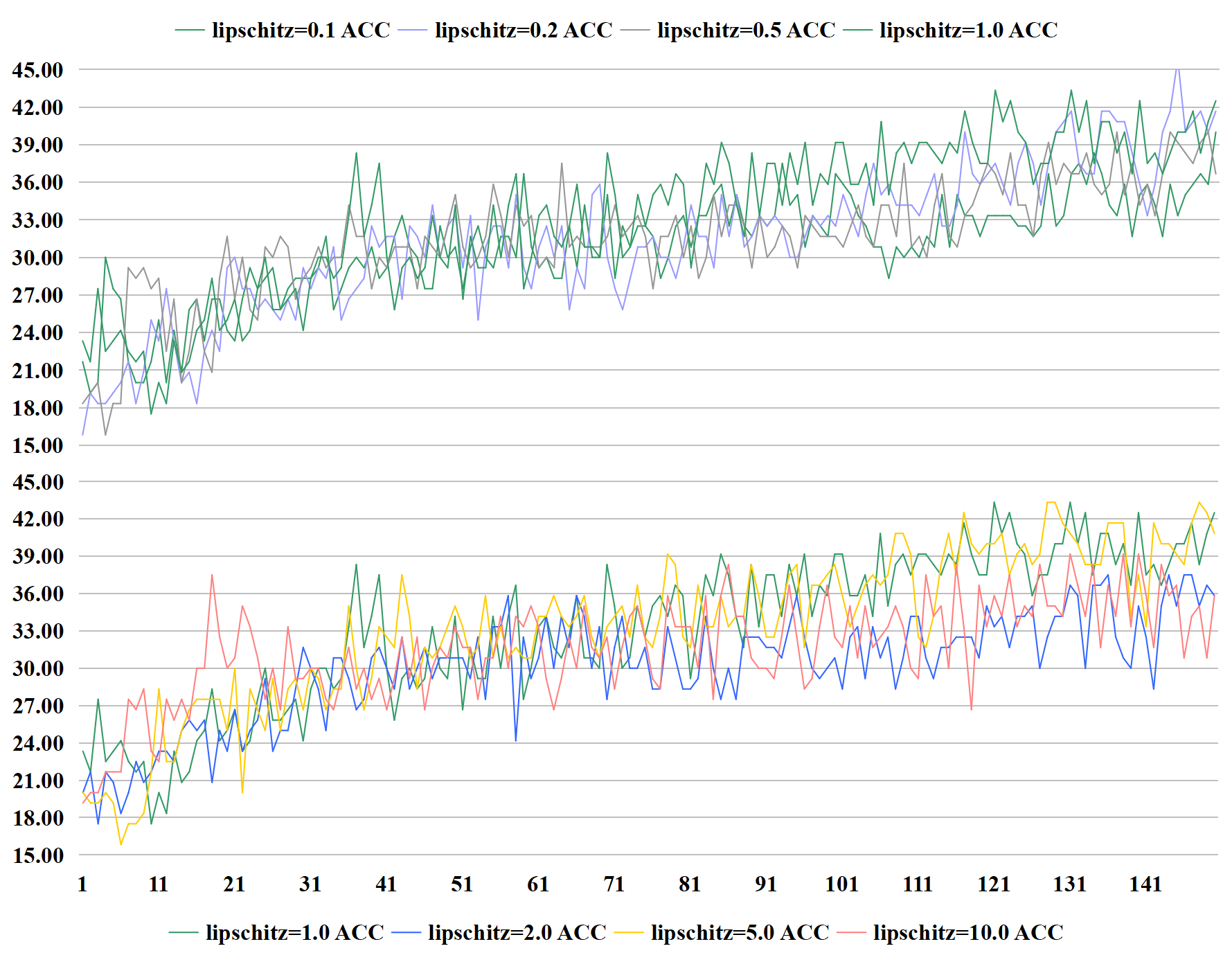}
    \caption{EAV Sub35 with different Lipschitz constant.}
    \label{fig:eav-lc2}
\end{figure}

\subsection{Supplementary Figure S3 for Analysis with ROC Curve}
\begin{figure}[H]
    \centering
    \includegraphics[width=0.8\textwidth]{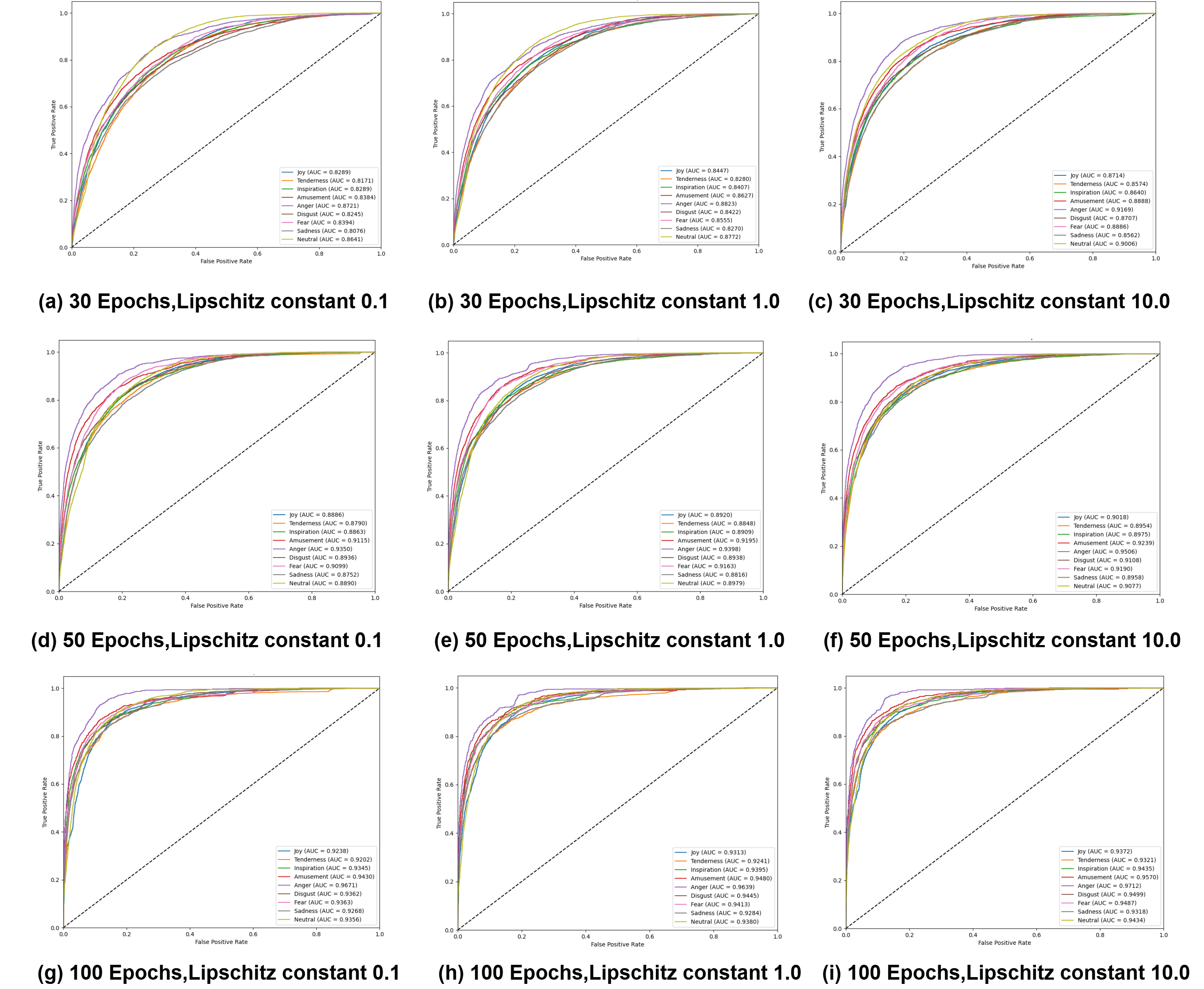}
    \caption{Nine ROC curves in a 3×3 grid, with Lipschitz constants 0.1, 1.0, and 10.0 from left to right, and training epochs 30, 50, and 100 from top to bottom. The curves show performance trade-offs across nine emotion categories.}
    \label{fig:roc}
\end{figure}

\subsection{Supplementary Figure S4 for Validation on the EAV Passive Dataset}
\begin{figure}[H]
    \centering
    \includegraphics[width=0.8\textwidth]{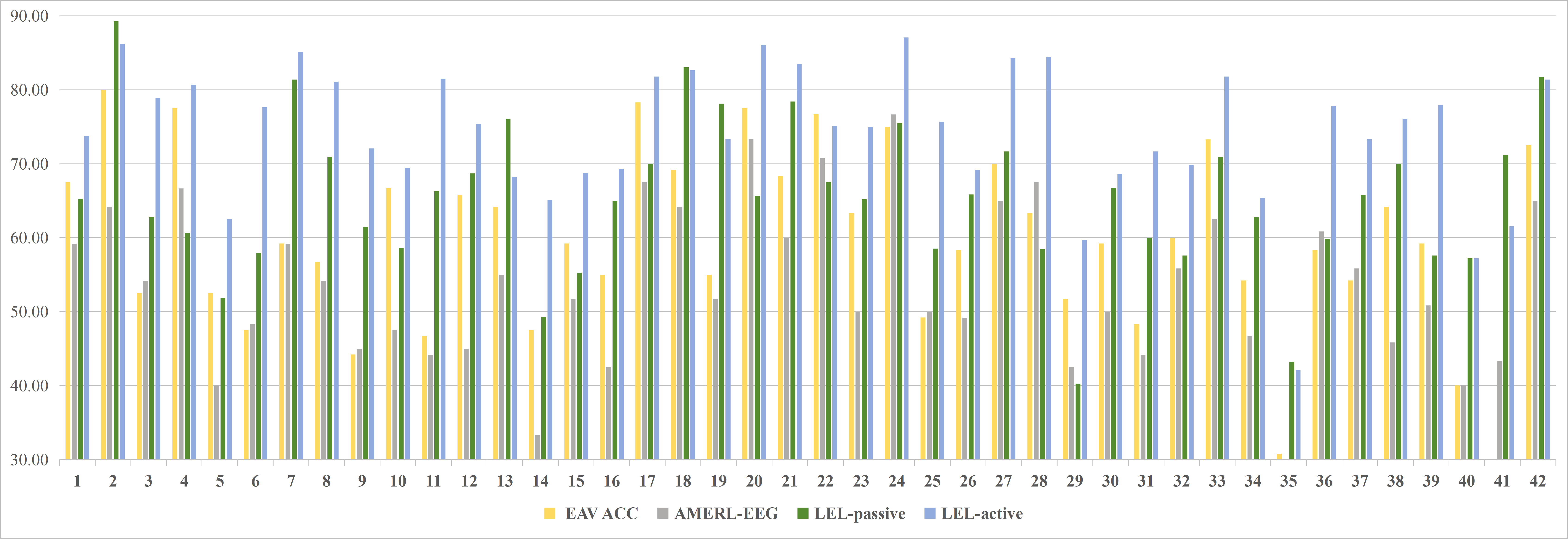}
    \caption{Using passive data experiments to compare with other active data experiments on the EAV dataset.}
    \label{fig:passive compare}
\end{figure}

\subsection{Supplementary Figure S5 for Correlation Between EEG Channels and Emotions}
\begin{figure}[H]
    \centering
    \includegraphics[width=0.7\textwidth]{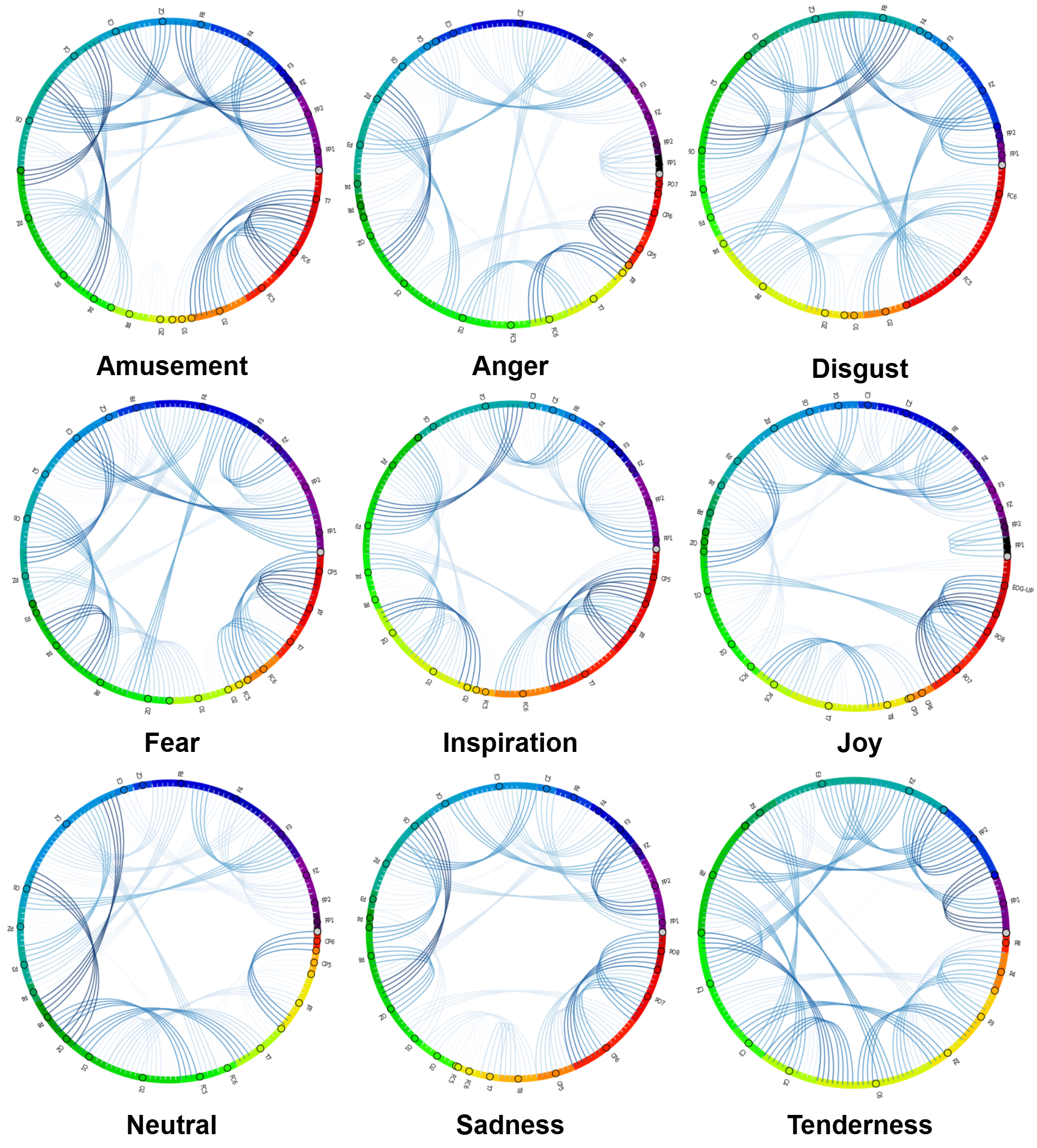}
    \caption{This figure shows nine chord diagrams representing different emotions. Each diagram illustrates the connections and relationships between various elements or features associated with the respective emotion.}
    \label{fig:channel-r}
\end{figure}

\section*{CRediT authorship contribution statement}
 
\textbf{Shengyu Gong:} Writing - original draft, Methodology, Formal analysis, Conceptualization, Visualization, Software; \textbf{Yueyang Li:} Writing - original draft, Investigation, Methodology; \textbf{Zijian Kang:} Validation, Data Curation, Investigation; \textbf{Bo Chai:} Validation, Writing - Review \& Editing; \textbf{Weiming Zeng:} Resources, Supervision, Writing - Review \& Editing; \textbf{Hongjie Yan:} Validation, Writing - Review \& Editing; \textbf{Zhiguo Zhang:} Validation, Writing - Review \& Editing; \textbf{Wai Ting Siok:} Funding acquisition, Validation, Writing - Review \& Editing; \textbf{Nizhuan Wang:} Conceptualization, Writing - review \& editing, Supervision, Project administration, Funding acquisition. 

\section*{Acknowledgments}
This work was supported by The Hong Kong Polytechnic University Start-up Fund (Project ID: P0053210), The Hong Kong Polytechnic University Faculty Reserve Fund (Project ID: P0053738), an internal grant from The Hong Kong Polytechnic University (Project ID: P0048377), The Hong Kong Polytechnic University Departmental Collaborative Research Fund (Project ID: P0056428), The Hong Kong Polytechnic University Collaborative Research with World-leading Research Groups Fund (Project ID: P0058097) and Research Grants Council Collaborative Research Fund (Project ID: P0049774).

\bibliographystyle{unsrt}
\bibliography{reference}

@inproceedings{Li2024,
  author = {Yuekang Li and Yidan Mao and Yifei Yang and Dongmian Zou},
  title = {Improving robustness of hyperbolic neural networks by Lipschitz analysis},
  booktitle = {Proceedings of the 30th ACM SIGKDD Conference on Knowledge Discovery and Data Mining},
  year = {2024},
  pages = {1713--1724},
}

@inproceedings{Dasoulas2021,
  title={Lipschitz normalization for self-attention layers with application to graph neural networks},
  author={Dasoulas, George and Scaman, Kevin and Virmaux, Aladin},
  booktitle={International Conference on Machine Learning},
  pages={2456--2466},
  year={2021}
}

@article{chaddad2023electroencephalography,
  title={Electroencephalography Signal Processing: {A} Comprehensive Review and Analysis of Methods and Techniques},
  author={Chaddad, Ahmad and Wu, Yihang and Kateb, Reem and Bouridane, Ahmed},
  journal={Sensors},
  volume={23},
  number={14},
  pages={6434},
  year={2023},
  publisher={MDPI}
}

@inproceedings{ACase-basedChannel,
author = {Du, Yang and Guan, Zijing and Huang, Weichen and Zhang, Xichun and Huang, Qiyun},
title = {A Case-based Channel Selection Method for {EEG} Emotion Recognition Using Interpretable Transformer Networks},
year = {2023},
isbn = {9798400708701},
publisher = {Association for Computing Machinery},
booktitle = {Proceedings of the 2023 International Conference on Computer, Vision and Intelligent Technology},
articleno = {3},
numpages = {5},
location = {Chenzhou, China},
series = {ICCVIT '23}
}

@inproceedings{ResearchonGCN-LSTM,
author = {Chang, Lina and Li, Qi and Yan, Xurong},
title = {Research on {GCN-LSTM} Emotion Recognition Algorithm with Attention Mechanism based on {EEG}},
year = {2024},
isbn = {9798400716645},
publisher = {Association for Computing Machinery},
booktitle = {Proceedings of the 2024 4th International Conference on Bioinformatics and Intelligent Computing},
pages = {170–174},
numpages = {5},
location = {Beijing, China},
series = {BIC '24}
}

@article{eav,
  title={{EAV: EEG-Audio-Video} Dataset for Emotion Recognition in Conversational Contexts},
  author={Lee, Min-Ho and Shomanov, Adai and Begim, Balgyn and Kabidenova, Zhuldyz and Nyssanbay, Aruna and Yazici, Adnan and Lee, Seong-Whan},
  journal={Scientific Data},
  volume={11},
  pages={1026},
  year={2024},
  publisher={Nature Publishing Group}
}

@article{faced,
  title={A Large Finer-grained Affective Computing {EEG} Dataset},
  author={Chen, Jingjing and Wang, Xiaobin and Huang, Chen and Hu, Xin and Shen, Xinke and Zhang, Dan},
  journal={Scientific Data},
  volume={10},
  number={740},
  year={2023},
  publisher={Nature Publishing Group}
}

@article{seed,
  title={Investigating critical frequency bands and channels for {EEG}-based emotion recognition with deep neural networks},
  author={Zheng, Wei-Long and Lu, Bao-Liang},
  journal={IEEE Transactions on Autonomous Mental Development},
  volume={7},
  number={3},
  pages={162--175},
  year={2015},
  publisher={IEEE}
}

@inproceedings{amerl,
  title={EEG-based Multimodal Representation Learning for Emotion Recognition},
  author={Yin, Kang and Shin, Hye-Bin and Li, Dan and Lee, Seong-Whan},
  booktitle={2025 13th International Conference on Brain-Computer Interface (BCI)},
  pages={1--4},
  year={2025},
  organization={IEEE}
}

@inproceedings{Jia2023,
  title={Enhancing node-level adversarial defenses by Lipschitz regularization of graph neural networks},
  author={Jia, Yaning and Zou, Dongmian and Wang, Hongfei and Jin, Hai},
  booktitle={Proceedings of the 29th ACM SIGKDD Conference on Knowledge Discovery and Data Mining},
  pages={951--963},
  year={2023}
}

@inproceedings{Ye2023,
  title={Mitigating transformer overconfidence via Lipschitz regularization},
  author={Ye, Wenqian and Ma, Yunsheng and Cao, Xu and Tang, Kun},
  booktitle={Uncertainty in Artificial Intelligence},
  pages={2422--2432},
  year={2023},
  organization={PMLR}
}

@article{hervas2023autism,
  title={Autism and depression: clinical presentation, evaluation and treatment},
  author={Herv{\'a}s, Amaia},
  journal={Medicina (Argentina)},
  volume={83},
  number={Suppl 2},
  pages={37--42},
  year={2023},
  publisher={Fundacion Revista Medicina (Buenos Aires)}
}

@article{houssien2022review,
  title={Human emotion recognition from {EEG-based} brain-computer interface using machine learning: a comprehensive review},
  author={Houssein, Essam H and Hammad, Asmaa and Ali, Abdelmgeid A},
  journal={Neural Computing and Applications},
  volume={34},
  number={15},
  pages={12527--12557},
  year={2022},
  publisher={Springer}
}

@article{li2022eeg,
  title={{EEG} Based Emotion Recognition: A Tutorial and Review},
  author={Li, Xiang and Zhang, Yazhou and Tiwari, Prayag and Song, Dawei and Hu, Bin and Yang, Meihong and Zhao, Zhigang and Kumar, Neeraj and Marttinen, Pekka},
  journal={ACM Computing Surveys},
  volume={55},
  number={4},
  pages={79:1--79:57},
  year={2022},
  publisher={ACM}
}

@article{zhao2025adaptive,
  title={Adaptive construction of critical brain functional networks for {EEG}-based emotion recognition},
  author={Zhao Ying and He Hong and Bi Xiaoying and Lu Yue},
  journal={Signal, Image and Video Processing},
  volume={19},
  number={1},
  pages={454},
  year={2025},
  publisher={Springer}
}

@article{geng2024deep,
  title={Deep learning-based {EEG} emotion recognition: a comprehensive review},
  author={Geng, Yuxiao and Shi, Shuo and Hao, Xiaoke},
  journal={Neural Computing and Applications},
  pages={1--32},
  year={2024},
  publisher={Springer}
}

@article{pillalamarri2025review,
  title={A review on {EEG-based} multimodal learning for emotion recognition},
  author={Pillalamarri, Rajasekhar and Shanmugam, Udhayakumar},
  journal={Artificial Intelligence Review},
  volume={58},
  number={5},
  pages={131},
  year={2025},
  publisher={Springer}
}

@article{kang2025hypergraph,
  title={Hypergraph Multi-Modal Learning for {EEG-Based} Emotion Recognition in Conversation},
  author={Kang, Zijian and Li, Yueyang and Gong, Shengyu and Zeng, Weiming and Yan, Hongjie and Bian, Lingbin and Siok, Wai Ting and Wang, Nizhuan},
  journal={arXiv preprint arXiv:2502.21154},
  year={2025}
}

@article{imtiaz2025towards,
  title={Towards Practical Emotion Recognition: {An} Unsupervised Source-Free Approach for {EEG} Domain Adaptation},
  author={Imtiaz, Md Niaz and Khan, Naimul},
  journal={arXiv preprint arXiv:2504.03707},
  year={2025}
}

@INPROCEEDINGS{li2024neural,
  author={Li, Yueyang and Kang, Zijian and Gong, Shengyu and Dong, Wenhao and Zeng, Weiming and Yan, Hongjie and Siok, Wai Ting and Wang, Nizhuan},
  booktitle={2025 IEEE International Conference on Multimedia and Expo (ICME)}, 
  title={Neural-MCRL: Neural Multimodal Contrastive Representation Learning for EEG-based Visual Decoding}, 
  year={2025},
  volume={},
  number={},
  pages={1-6},
  doi={10.1109/ICME59968.2025.11210130}}

@article{zhang2024mini,
  title={Mini review: Challenges in {EEG} emotion recognition},
  author={Zhang, Zhihui and Fort, Josep M and Gim{\'e}nez Mateu, Lluis},
  journal={Frontiers in Psychology},
  volume={14},
  pages={1289816},
  year={2024},
  publisher={Frontiers Media SA}
}

@article{rashid2020current,
  title={Current status, challenges, and possible solutions of {EEG-based} brain-computer interface: a comprehensive review},
  author={Rashid, Mamunur and Sulaiman, Norizam and PP Abdul Majeed, Anwar and Musa, Rabiu Muazu and Ab. Nasir, Ahmad Fakhri and Bari, Bifta Sama and Khatun, Sabira},
  journal={Frontiers in Neurorobotics},
  volume={14},
  pages={25},
  year={2020},
  publisher={Frontiers Media SA}
}

@article{fiorini2024eeg,
  title={EEG-based emotional valence and emotion regulation classification: a data-centric and explainable approach},
  author={Fiorini, Linda and Bossi, Francesco and Di Gruttola, Francesco},
  journal={Scientific Reports},
  volume={14},
  number={1},
  pages={24046},
  year={2024},
  publisher={Nature Publishing Group UK London}
}

@article{zhong2020eeg,
  title={EEG-based emotion recognition using regularized graph neural networks},
  author={Zhong, Peixiang and Wang, Di and Miao, Chunyan},
  journal={IEEE Transactions on Affective Computing},
  volume={13},
  number={3},
  pages={1290--1301},
  year={2020},
  publisher={IEEE}
}

@inproceedings{jia2024aligning,
  title={Aligning relational learning with lipschitz fairness},
  author={Jia, Yaning and Zhang, Chunhui and Vosoughi, Soroush},
  booktitle={The Twelfth International Conference on Learning Representations},
  year={2024}
}

@article{ResearchProgressofEEG-Based,
  title={Research progress of {EEG-based} emotion recognition: a survey},
  author={Wang, Yiming and Zhang, Bin and Di, Lamei},
  journal={ACM Computing Surveys},
  volume={56},
  number={11},
  pages={1--49},
  year={2024},
  publisher={ACM New York, NY}
}

@inproceedings{Spatial-temporalTransformers,
  title={Spatial-temporal transformers for {EEG} emotion recognition},
  author={Liu, Jiyao and Wu, Hao and Zhang, Li and Zhao, Yanxi},
  booktitle={Proceedings of the 6th International Conference on Advances in Artificial Intelligence},
  pages={116--120},
  year={2022}
}

@inproceedings{Researchontwo-dimensional,
  title={Research on two-dimensional convolution-based EEG emotion recognition methods},
  author={Li, Hongna and Li, Yue},
  booktitle={Proceedings of the 2024 International Conference on Biomedicine and Intelligent Technology},
  pages={147--151},
  year={2024}
}

@article{10947211,
  author={Li, Yueyang and Zeng, Weiming and Dong, Wenhao and Han, Di and Chen, Lei and Chen, Hongyu and Kang, Zijian and Gong, Shengyu and Yan, Hongjie and Siok, Wai Ting and Wang, Nizhuan},
  journal={IEEE Transactions on Instrumentation and Measurement}, 
  title={A Tale of Single-Channel Electroencephalography: Devices, Datasets, Signal Processing, Applications, and Future Directions}, 
  year={2025},
  volume={74},
  pages={1-20},
}

@article{li2025freqdgt,
  title={{FreqDGT}: Frequency-Adaptive Dynamic Graph Networks with Transformer for Cross-subject {EEG} Emotion Recognition},
  author={Li, Yueyang and Gong, Shengyu and Zeng, Weiming and Wang, Nizhuan and Siok, Wai Ting},
  journal={arXiv preprint arXiv:2506.22807},
  year={2025}
}

@article{chen2025eeg,
  title={{EEG} emotion copilot: Optimizing lightweight {LLMs} for emotional {EEG} interpretation with assisted medical record generation},
  author={Chen, Hongyu and Zeng, Weiming and Chen, Chengcheng and Cai, Luhui and Wang, Fei and Shi, Yuhu and Wang, Lei and Zhang, Wei and Li, Yueyang and Yan, Hongjie and others},
  journal={Neural Networks},
  volume={192},
  pages={107848},
  year={2025},
  publisher={Elsevier}
}

@article{feng2024neural,
  title={Neural modulation alteration to positive and negative emotions in depressed patients: Insights from fmri using positive/negative emotion atlas},
  author={Feng, Yu and Zeng, Weiming and Xie, Yifan and Chen, Hongyu and Wang, Lei and Wang, Yingying and Yan, Hongjie and Zhang, Kaile and Tao, Ran and Siok, Wai Ting and others},
  journal={Tomography},
  volume={10},
  number={12},
  pages={2014--2037},
  year={2024},
  publisher={MDPI}
}

@article{zhang2024stanet,
  title={{STANet}: {A} novel spatio-temporal aggregation network for depression classification with small and unbalanced {FMRI} data},
  author={Zhang, Wei and Zeng, Weiming and Chen, Hongyu and Liu, Jie and Yan, Hongjie and Zhang, Kaile and Tao, Ran and Siok, Wai Ting and Wang, Nizhuan},
  journal={Tomography},
  volume={10},
  number={12},
  pages={1895--1914},
  year={2024},
  publisher={MDPI}
}

@article{cai2025mm,
  title={{MM-GTUNets}: {Unified} multi-modal graph deep learning for brain disorders prediction},
  author={Cai, Luhui and Zeng, Weiming and Chen, Hongyu and Zhang, Hua and Li, Yueyang and Feng, Yu and Yan, Hongjie and Bian, Lingbin and Siok, Wai Ting and Wang, Nizhuan},
  journal={IEEE Transactions on Medical Imaging},
  volume={44},
  number={9},
  pages={3705--3716},
  year={2025},
  publisher={IEEE}
}

@article{li2025information,
  title={Information Bottleneck-Guided Heterogeneous Graph Learning for Interpretable Neurodevelopmental Disorder Diagnosis},
  author={Li, Yueyang and Chen, Lei and Dong, Wenhao and Gong, Shengyu and Kang, Zijian and Wei, Boyang and Zeng, Weiming and Yan, Hongjie and Bian, Lingbin and Siok, Wai Ting and others},
  journal={arXiv preprint arXiv:2502.20769},
  year={2025}
}

@article{li2025mhnet,
  title={{MHNet}: {Multi-view} high-order network for diagnosing neurodevelopmental disorders using resting-state {fMRI}},
  author={Li, Yueyang and Zeng, Weiming and Dong, Wenhao and Cai, Luhui and Wang, Lei and Chen, Hongyu and Yan, Hongjie and Bian, Lingbin and Wang, Nizhuan},
  journal={Journal of Imaging Informatics in Medicine},
  volume={38},
  pages={2994--3014},
  year={2025},
  publisher={Springer}
}

@article{dong2025starformer,
  title={{STARFormer}: {A} Novel Spatio-Temporal Aggregation Reorganization Transformer of {FMRI} for Brain Disorder Diagnosis},
  author={Dong, Wenhao and Li, Yueyang and Zeng, Weiming and Chen, Lei and Yan, Hongjie and Siok, Wai Ting and Wang, Nizhuan},
  journal={Neural Networks},
  volume={192},
  pages={107927},
  year={2025},
  publisher={Elsevier}
}

@article{hirsch2018emotional,
  title={Emotional dysregulation is a primary symptom in adult {Attention-Deficit/Hyperactivity Disorder (ADHD)}},
  author={Hirsch, Oliver and Chavanon, MiraLynn and Riechmann, Elke and Christiansen, Hanna},
  journal={Journal of Affective Disorders},
  volume={232},
  pages={41--47},
  year={2018},
  publisher={Elsevier}
}

@article{conner2021emotion,
  title={Emotion dysregulation is substantially elevated in autism compared to the general population: {Impact} on psychiatric services},
  author={Conner, Caitlin M and Golt, Josh and Shaffer, Rebecca and Righi, Giulia and Siegel, Matthew and Mazefsky, Carla A},
  journal={Autism Research},
  volume={14},
  number={1},
  pages={169--181},
  year={2021},
  publisher={Wiley Online Library}
}

@article{joormann2016examining,
  title={Examining emotion regulation in depression: {A} review and future directions},
  author={Joormann, Jutta and Stanton, Colin H},
  journal={Behaviour Research and Therapy},
  volume={86},
  pages={35--49},
  year={2016},
  publisher={Elsevier}
}

@article{kimhy2012emotion,
  title={Emotion awareness and regulation in individuals with schizophrenia: Implications for social functioning},
  author={Kimhy, David and Vakhrusheva, Julia and Jobson-Ahmed, Lauren and Tarrier, Nicholas and Malaspina, Dolores and Gross, James J},
  journal={Psychiatry Research},
  volume={200},
  number={2-3},
  pages={193--201},
  year={2012},
  publisher={Elsevier}
}

@article{ACCNet,
  title={{ACCNet}: {Adaptive} Cross-frequency Coupling Graph Attention for {EEG} Emotion Recognition},
  author={Tian, Dongyuan and Wang, Yucheng and Gong, Peiliang and Xu, Zhewen and Chen, Zhenghua and Wei, Xiaohui and Wu, Min},
  journal={Neural Networks},
  volume={191},
  pages={107853},
  year={2025},
  publisher={Elsevier}
}

@inproceedings{fcstgnn,
  title={Fully-connected spatial-temporal graph for multivariate time-series data},
  author={Wang, Yucheng and Xu, Yuecong and Yang, Jianfei and Wu, Min and Li, Xiaoli and Xie, Lihua and Chen, Zhenghua},
  booktitle={Proceedings of the AAAI Conference on Artificial Intelligence},
  volume={38},
  number={14},
  pages={15715--15724},
  year={2024}
}

@article{eegnet,
  title={{EEGNet}: a compact convolutional neural network for {EEG-based} brain-computer interfaces},
  author={Lawhern, Vernon J and Solon, Amelia J and Waytowich, Nicholas R and Gordon, Stephen M and Hung, Chou P and Lance, Brent J},
  journal={Journal of Neural Engineering},
  volume={15},
  number={5},
  pages={056013},
  year={2018},
  publisher={iOP Publishing}
}

@article{lggnet,
  title={{LGGNet}: {Learning} from local-global-graph representations for brain-computer interface},
  author={Ding, Yi and Robinson, Neethu and Tong, Chengxuan and Zeng, Qiuhao and Guan, Cuntai},
  journal={IEEE Transactions on Neural Networks and Learning Systems},
  volume={35},
  number={7},
  pages={9773--9786},
  year={2023},
  publisher={IEEE}
}

@article{bfgcn,
  author={Li, Cunbo and Tang, Tian and Pan, Yue and Yang, Lei and Zhang, Shuhan and Chen, Zhaojin and Li, Peiyang and Gao, Dongrui and Chen, Huafu and Li, Fali and Yao, Dezhong and Cao, Zehong and Xu, Peng},
  journal={IEEE Transactions on Neural Networks and Learning Systems}, 
  title={An Efficient Graph Learning System for Emotion Recognition Inspired by the Cognitive Prior Graph of {EEG} Brain Network}, 
  year={2025},
  volume={36},
  number={4},
  pages={7130-7144}
}

@inproceedings{fbcnet,
  title={A multi-view {CNN} with novel variance layer for motor imagery brain computer interface},
  author={Mane, Ravikiran and Robinson, Neethu and Vinod, A Prasad and Lee, Seong-Whan and Guan, Cuntai},
  booktitle={2020 42nd Annual International Conference of the IEEE Engineering in Medicine \& Biology Society (EMBC)},
  pages={2950--2953},
  year={2020},
  organization={IEEE}
}

@ARTICLE{11262248,
  author={Li, Hanyu and Kim, Byung Hyung},
  journal={IEEE Transactions on Instrumentation and Measurement}, 
  title={Hierarchical Dynamic Local–Global-Graph Representation Learning for EEG Emotion Recognition}, 
  year={2025},
  volume={74},
  number={},
  pages={1-14},
  doi={10.1109/TIM.2025.3635321}}

@ARTICLE{10960695,
  author={Ding, Yi and Tong, Chengxuan and Zhang, Shuailei and Jiang, Muyun and Li, Yong and Lim, Kevin JunLiang and Guan, Cuntai},
  journal={IEEE Transactions on Neural Networks and Learning Systems}, 
  title={EmT: A Novel Transformer for Generalized Cross-Subject EEG Emotion Recognition}, 
  year={2025},
  volume={36},
  number={6},
  pages={10381-10393},
  doi={10.1109/TNNLS.2025.3552603}}

@article{Chen_2025,
doi = {10.1088/1741-2552/ade290},
year = {2025},
month = {jun},
publisher = {IOP Publishing},
volume = {22},
number = {3},
pages = {031004},
author = {Chen, Yuxin and Peng, Yong and Tang, Jiajia and Camilleri, Tracey and Camilleri, Kenneth and Kong, Wanzeng and Cichocki, Andrzej},
title = {EEG-based affective brain–computer interfaces: recent advancements and future challenges},
journal = {Journal of Neural Engineering}
}

@ARTICLE{9817639,
  author={Peng, Yong and Wang, Wenjuan and Kong, Wanzeng and Nie, Feiping and Lu, Bao-Liang and Cichocki, Andrzej},
  journal={IEEE Transactions on Affective Computing}, 
  title={Joint Feature Adaptation and Graph Adaptive Label Propagation for Cross-Subject Emotion Recognition From EEG Signals}, 
  year={2022},
  volume={13},
  number={4},
  pages={1941-1958},
  doi={10.1109/TAFFC.2022.3189222}}

@article{KHARE2024102019,
title = {Emotion recognition and artificial intelligence: A systematic review (2014–2023) and research recommendations},
journal = {Information Fusion},
volume = {102},
pages = {102019},
year = {2024},
issn = {1566-2535},
doi = {https://doi.org/10.1016/j.inffus.2023.102019},
author = {Smith K. Khare and Victoria Blanes-Vidal and Esmaeil S. Nadimi and U. Rajendra Acharya}
}

@article{ZHANG2024121692,
title = {Deep learning-based multimodal emotion recognition from audio, visual, and text modalities: A systematic review of recent advancements and future prospects},
journal = {Expert Systems with Applications},
volume = {237},
pages = {121692},
year = {2024},
issn = {0957-4174},
doi = {https://doi.org/10.1016/j.eswa.2023.121692},
author = {Shiqing Zhang and Yijiao Yang and Chen Chen and Xingnan Zhang and Qingming Leng and Xiaoming Zhao}
}

@article{zhang2024application,
  author    = {Zhang, Min and Yang, Yi and Zhao, Yongmei and Sui, Changbai and Sui, Ying and Jiang, Youzhi and Liu, Kanlai and Yang, Shuai and Wang, Liqin and Chen, Bingjie and Zhang, Rui and Zhang, Qun and Huang, Zhisheng and Huang, Manli},
  title     = {The application of integrating electroencephalograph-based emotion recognition technology into brain--computer interface systems for the treatment of depression: a narrative review},
  journal   = {Advanced Technology in Neuroscience},
  year      = {2024},
  volume    = {1},
  number    = {2},
  pages     = {188--200},
  month     = {dec},
  doi       = {10.4103/ATN.ATN-D-24-00018}
}

@article{dogra2025development,
  author    = {Dogra, Saanvi and Kouznetsova, Valentina L. and Kesari, Santosh and Tsigelny, Igor F.},
  title     = {Development of a miRNA-based deep learning model for autism spectrum disorder diagnosis},
  journal   = {Advanced Technology in Neuroscience},
  year      = {2025},
  volume    = {2},
  number    = {2},
  pages     = {72--76},
  month     = {jun},
  doi       = {10.4103/ATN.ATN-D-24-00033}
}

\end{document}